\documentclass[10pt]{cai26}

\usepackage{algorithm}
\usepackage{algpseudocode}

\usepackage{microtype}
\usepackage{graphicx}
\usepackage{booktabs}
\usepackage{tabularx}
\usepackage{multirow}
\usepackage{longtable}

\usepackage{amsmath}
\usepackage{amssymb}
\usepackage{mathtools}
\usepackage{amsthm}

\begin{document}
\def\conferenceyear{2026}
\volumeheader{39}{0}
\begin{center}

\title{ADAPTive Input Training for Many-to-One Pre-Training on Time-Series Classification}

\maketitle

\thispagestyle{empty}
\pagenumbering{gobble}

\begin{tabular}{cc}
Paul Quinlan\upstairs{\affilone,\affilthree,*}, Qingguo Li\upstairs{\affiltwo,\affilthree}, Xiaodan Zhu\upstairs{\affilone,\affilthree}
\\[0.25ex]
{\small \upstairs{\affilone} Electrical and Computer Engineering, Queen's University} \\
{\small \upstairs{\affiltwo} Mechanical and Materials Engineering, Queen's University} \\
{\small \upstairs{\affilthree} Ingenuity Labs Research Institute, Queen's University} \\
\end{tabular}
  
\emails{
  \upstairs{*} 15pwq@queensu.ca 
}
\vspace*{0.1in}
\end{center}

\begin{abstract} 
Recent work on time-series models has leveraged self-supervised training to learn meaningful features and patterns in order to improve performance on downstream tasks and generalize to unseen modalities. While these pretraining methods have shown great promise in one-to-many scenarios, where a model is pre-trained on one dataset and fine-tuned on a downstream dataset, they have struggled to generalize to new datasets when more datasets are added during pre-training. This is a fundamental challenge in building foundation models for time-series data, as it limits the ability to develop models that can learn from a large variety of diverse datasets available. To address this challenge, we present a new pre-training paradigm for time-series data called \texttt{ADAPT}, which can efficiently align the physical properties of data in the time-series domain, enabling mixed-batch pre-training despite the extreme discrepancies in the input sizes and channel dimensions of pre-training data. We trained on 162 time-series classification datasets and set new state-of-the-art performance for classification benchmarks. We successfully train a model within the time-series domain on a wide range of datasets simultaneously, which is a major building block for building generalist foundation models in time-series domains.
\end{abstract}

\copyrightnotice

\section{Introduction}

Analysis of time-series data is critical in many real-life applications, including those in the medical~\cite{Acosta2022}, financial ~\cite{10.1145/3502289}, industrial~\cite{app132212374}, agricultural~\cite{Amankulova2023} and environmental domains~\cite{amoatey2024effects}, among others ~\cite{LIU2023501,9745049}. Pre-training and transfer learning have allowed for the application of large models to diverse tasks even when there is limited task-specific data available. However, the unique characteristics and variability of time-series data often make it challenging to develop generalist models that can be successfully adapted to different downstream modalities after pretraining. Previous research has mainly focused on custom modality-specific designs, pre-trained on a singular dataset, to improve the inductive bias of model training. This approach is in contrast to the dominant strategies used in other domains, such as natural language processing (NLP) and computer vision (CV), which focus on training a single model on many large datasets and has led to the creation of \textit{foundation models} such as GPT4 ~\cite{openai2023gpt4}, Mixtral \cite{jiang2024mixtral}, LaMDA~\cite{thoppilan2022lamda}, wav2vec2.0 ~\cite{10.5555/3495724.3496768}, DALLE-2 ~\cite{ramesh2022hierarchical}, T5 ~\cite{JMLR:v21:20-074}, among others.

These models have shown remarkable performance and ability to generalize to unseen data and tasks. While pre-trained models have been applied to differing downstream modalities via fine-tuning (one-to-many), no work has been able to successfully pre-train a time-series model over a wide range of time-series types and modalities. The work performed in ~\cite{zhang2022selfsupervised} found that adding datasets during pre-training inhibited learning and rapidly degrade model performance (a roughly 25\% decrease in performance between the one-to-one and four-to-one scenario). Each of the above foundation models illustrates a relatively simple correlation between volume of data, model size, and performance; by training larger models on large datasets, we can train powerful models with exceptional transfer learning capabilities. 

Many-to-one training is very desirable as it may allow us to scale up the amount of training data, model size, ease transferring to new domains and modalities and allow us to train one large model for application on unseen time-series. Furthermore, there are many negative future implications in failing to pre-train models in a many-to-one or many-to-many setting. Refer to the Analysis and Limitations section for more details.

To address this challenge, we propose a new framework that achieves state-of-the-art performance on time-series classification benchmarks. Our framework trains a single model on 162 time-series datasets for classification, each with varying length, channel dimension, and modality. To create an input-agnostic model, we propose the use of average adaptive pooling during the data loading process, which facilitates mixed batch training for time-series data. Our framework is also designed to be completely model agnostic, which allows us to leverage future improvements in time-series-specific model architectures.
Our proposed framework addresses several key challenges in building foundation models for time-series analysis: 
(i) The models must be able to process inputs of any dimension, modality, or channel sizes.  
(ii) They need to be trained using modern parallel computing strategies. Primarily we would like to train the models with large batch sizes in mixed batch training, requiring all data be aligned in a way that it can be mixed within batches with the same input length and modality size. 
(iii) A great deal of research has gone into creating powerful time-series specific model architectures. We would like our training strategy to be completely model agnostic to leverage possible future improvements in this research area.


By introducing the \texttt{ADAPT} framework, we brought forward the state-of-the-art performance of pretrained models on diverse time-series sensor data, which will in turn contribute to various real-life downstream tasks.  


\section{Related Work}
\label{sec:related}
\paragraph{Foundation Models in Time-Series.} Foundation models have been trained on vast datasets for broad applicability across various tasks, as defined in \cite{Stanford_CRFM_2021}. However, some models, either in full or in part, claim this designation in the time-series domain, such as those mentioned in recent studies~\cite{zhou2023fits,wu2023timesnet}, do not meet the CRFM's criteria, not just because they lack the extensive training datasets but also the appropriate pretrianing architectures that can help counteract and solve the problem. These models' primary limitation is the relative scarcity of training data. The availability of diverse time-series data suggests that adopting a many-to-one pre-training approach may overcome this hurdle, enabling effective adaptation to different time-series applications.

\paragraph{Pre-Training Strategies.}
Foundation models have gained a significant status in the fields of NLP and computer vision, exhibiting remarkable success in solving a wide range of problems~\cite{zhou2023comprehensive}. These models largely owe their success to self-supervised pretraining strategies that utilize modern parallel computing techniques to train on massive amounts of data. Recent research, such as ~\cite{touvron2023llama}, has found that, besides advancements in model architecture, training on more data for longer periods is critical in building powerful foundation models. Adding in more datasets for these models is simple since the underlying data in each of the commonly studied domains (i.e., text or image) is consistent across datasets.

Some recent works have focused on skipping the pre-training phase of time-series models by simply adopting pre-trained models from other domains~\cite{zhou2023fits,li2023time,chang2024llm4ts}. The purpose is to take large pre-trained language or vision models and finetune them for downstream tasks. While we think this work is important, it still does not provide us with a solution for training and scaling models on diverse time-series. Investigating time-series specific pre-training strategies play a fundamental role in both building time-series specific models but also as components in future multi-modality time-series models, for example in time-series specific variants of CLIP~\cite{radford2021learning}.

Time-series are unique in terms of modality types, input lengths, modality dimensions, sampling rates, and other factors. As a result, recent works have focused on learning universal features for the time-series domain that transfer between time-series ~\cite{Yue_Wang_Duan_Yang_Huang_Tong_Xu_2022},\cite{zhang2022selfsupervised}. One common feature of these works is the emphasis on learning strategies in which the model only has access to one dataset during pretraining and is fine-tuned on downstream domain data. Since there is no effective solution for aligning different data during pretraining, this has been a realistic scenario as the time-series domain comprises of a wide range of data modalities with varying characteristics. 
Some recent examples of self-supervised frameworks for pre-training within the time-series domain include ts2vec~\cite{Yue_Wang_Duan_Yang_Huang_Tong_Xu_2022}, TS-TCC~\cite{ijcai2021p324}, COST \cite{woo2022cost}, CLOCS~\cite{Kiyasseh2020CLOCSCL}, CLUDA \cite{ozyurt2023contrastive} and TNC~\cite{tonekaboni2021unsupervised}. Each of these models represents important contributions to acquiring universal representations from time-series datasets in the \textit{one-to-many} scenario using self-supervised learning.

\paragraph{Many-to-one Pre-training.}
Previous work in training models in the many-to-one setting for time-series is limited, only explored very recently in TF-C~\cite{zhang2022selfsupervised}. This method focuses on a contrastive objective between the time and frequency components of the input signal to improve generalization from the pretraining dataset to the target modality. TF-C sets a state-of-the-art performance for time-series classification; however when trained in the many-to-one scenario, model performance degrades rapidly. In order to allow for many-to-one training they merge four pre-training datasets into one (SleepEEG, FD-A, HAR, and ECG). To address mismatch in dimensionalities, they limited each dataset to only one channel dimension and truncated or padded each dataset 1,500 observations. Their experiment demonstrated a 25\% decrease in performance compared to pre-training on a singular dataset.

A comparable solution for implementing mixed-batch training across multiple modalities is presented in GATO~\cite{reed2022generalist}, which is a large foundation model trained on controls, text, and image data. One key contribution is their embedding scheme which allows for proprioception data, image, text, continuous actions and discrete agent actions to be represented in a mixed batch form. They performed this by applying a separate embedding structure for each modality and each embedding strategy matches the channel dimensions between the modalities. While it may seem like this structure could work for the time-series domain, there are far too many modalities to adapt this strategy (in our setup, it would require over 150 different embedding strategies). Secondly, in the time-series domain, embedding is incorporated into the structure of the model (usually as linear or convolutional layers), meaning that in order to separate the model and embedding, we would first need to train a small embedding model for each dataset in the database. Finally, while the relative lengths of the input sequences in GATO were similar, data within the time-series domain significantly differs in input length, meaning that aligning the channel dimensions is insufficient. Our proposed solution, \texttt{ADAPT}, overcomes all these challenges and provides a basis for helping construct foundation models in the time-series domains.

\section{Our Approach}

\subsection{Model Overview}
Within the time-series domains, it has been standard practice to pre-train models using only a singular type of data and fine-tune them on a specific target task, which we call the \textit{one-to-one} setup. In general,
the models for diverse time series yet struggle to learn high-quality and transferable representations, under the \textit{many-to-one} setting (refer to the related work section for more discussion),  when the properties of the pretraining database are diverse or heterogeneous. 
Unlike the well-studied models that focus on homogeneous data (e.g., the large language models), our approach targets the time-series data that are made up of many diverse modalities and from different sources, such as smart home, human motion, healthcare, and fault detection data. The diverse time-series data have varying physical properties and channel dimensions, and no research yet has found an efficient way of unifying these properties for pretraining. 
The lack of a universal method for performing alignment across datasets prevents models in the time-series domain from building better representations. 

\begin{figure*}[t]
  \centering
  \includegraphics[width=0.99\textwidth]{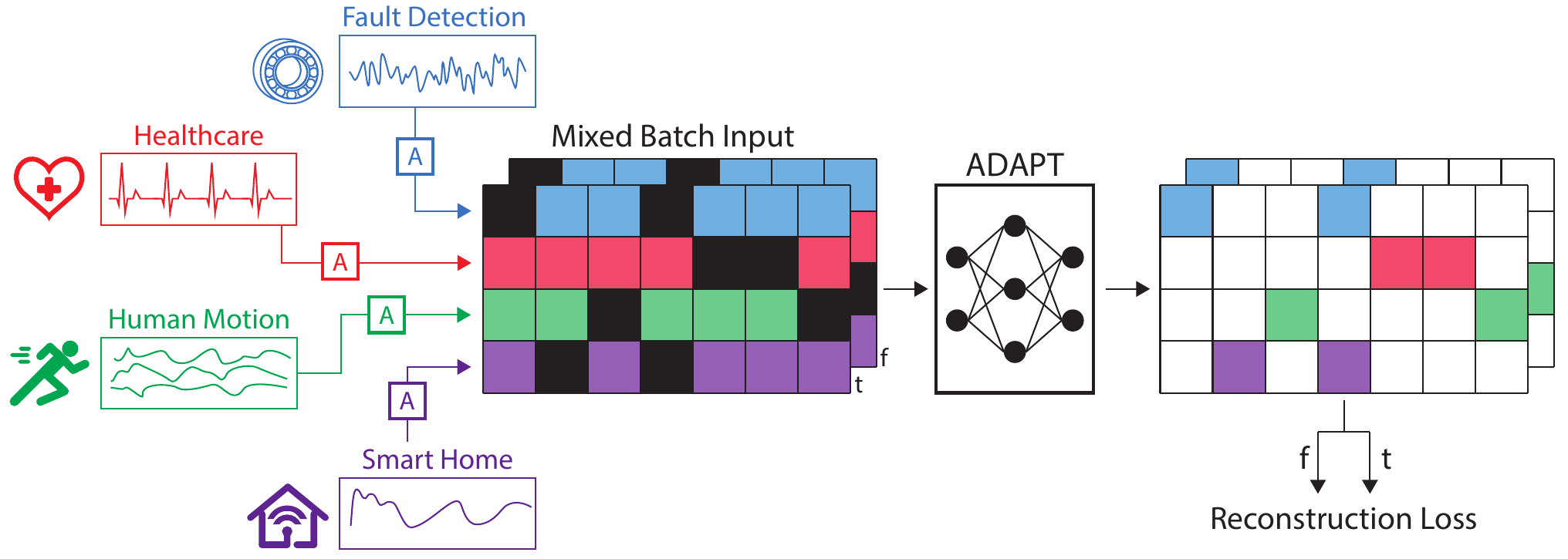}
  \caption{An overview of the adapt process and time-frequency masking algorithm. Each time-series is adaptive-pooled (in the [A] boxes) to a universal representation space and enable mixed-batch training among different modality types. At training time we add noise to the joint representation space in both the time and frequency space.}
  \label{fig:ADAPT}
\end{figure*}

The overall architecture of the proposed \texttt{ADAPT} approach is shown in Figure \ref{fig:ADAPT}, which aims to design an effective method for building a unified representation space in the many-to-one training scenario by aligning physical properties for training and enabling different combination of time-series datasets and modalities during pretraining and down-stream finetuning.
We believe that domain shift and adaptation can be mitigated by training a model on large volumes of increasingly heterogeneous data. If successful, transferring knowledge between domains should be easier as the model will have likely seen some related data during training.  

\subsection{Adaptive Input Training}
Time-Series data consists of widely different input lengths and channel dimensions. This means that truncation or padding will may seriously degrade or dilute the inputs when there are large discrepancies between the maximum and minimum input size. Difficulties in aligning channel dimensions also put constraints on the model architecture requirements for processing varying channel dimensions. The physical constraints mean that allowing mixed batch training for time-series based models without seriously degrading model performance was an open problem. We propose the use of average adaptive pooling applied during data-loading in order to transform each input to the same representation space regardless of size. 

We recognize that adaptively pooling the data risks losing fine-grained information in the data and may cause temporal distortions. However, the dimensional size of many popular datasets and data types are often smaller than that of the target dimension for embedding. This means that we are often up-sampling the data. We will show that any accuracy losses caused by the temporal distortions are mitigated by the increase in model performance using \texttt{ADAPT}.

Each input is adaptively pooled before batching to enable mixed batch pre-training. Refer to Algorithm \ref{alg:adaptivepool} for a summary of our data processing strategy. While adaptive pooling could be applied for any pre-training strategy and any model architecture, the stability of the model is not guaranteed. We have created a masked-language-modeling-inspired pretraining strategy using span-masking which also considers the frequency components of the input signal. Both the time and frequency input domains are embedded via separate time and frequency encoders and then fed to a stacked transformer encoder model with two separate MLP layers applied to the output to provide reconstruction loss in the time and frequency domain. Motivated by ~\cite{wagh2022evaluating}, which found that noise can emulate realistic data shifts in EEG data, we add random Gaussian noise to each sample during training.

\begin{algorithm}[h]
\caption{Data processing for diverse time series. Data is loaded, normalized, and converted to frequency representation via FFT before adaptive pooling. The processed data from all datasets is then combined. Samples from the combined data undergo noise addition (ADDNOISE) and span-masking (SPANMASK) for batching.}
\label{alg:adaptivepool}
\begin{algorithmic}[1]
\State Initialize $TrainingDataset \gets \emptyset$
\For{each $dataset$}
    \State $data[time] \gets \text{LoadData}(dataset)$
    \State $data[freq] \gets \text{FFT}(data[time])$
    \State \text{AdaptPool}(data[time], data[freq])
    \State Append $data$ to $TrainingDataset$
\EndFor
\While{$length(batch) < batchSize$}
    \State $sample \gets$ Extract sample from $TrainingDataset$
    \State $sample \gets \text{AddNoise}(sample)$
    \State $sample \gets \text{SpanMask}(sample)$
    \State Add current $sample$ to $batch$
\EndWhile
\end{algorithmic}
\end{algorithm}

\begin{algorithm}
\caption{Adaptive Pooling Procedure}\label{alg:cap}
\label{alg:adaptivepool}
   \begin{algorithmic}[1]
    \State $x \gets$ input, output, $D$, $H_{in}$, $W_{in}$, $H_{out}$, $W_{out}$,
       Stride$_D$, Stride$_H$, Stride$_W$

    \Function{$Start\char`_Index$}{$idx,O,I$}
    \State \Return $floor((idx*I)/O)$
    \EndFunction
    \Function{$End\char`_Index$}{$idx,O,I$}
    \State \Return $ceil(((idx+1)*I)/O)$
    \EndFunction
    \Function{AverageAdaptivePool2d}{$x$}
        \For{$d \gets 0$ to $D$}
            \For{$h \gets 0$ to $H_{out}$}
                \State $H_{start} =Start\char`_Index(h,H_{out},H_{in})$
                \State $H_{end} =End\char`_Index(h,H_{out},H_{in})$
                \State $k_H = H_{end}-H_{start}$
                \For{$w \gets 0$ to $W_{out}$}
                    \State $W_{start} =Start\char`_Index(w,W_{out},W_{in})$
                    \State $W_{end} =End\char`_Index(w,W_{out},W_{in})$
                    \State $k_W = W_{end}-W_{start}$
                    \State $output[d,h,w] = mean(input[d,h:h+k_H,w+k_W]) $
        \EndFor
        \EndFor
        \EndFor
    \EndFunction
  \end{algorithmic}
\end{algorithm} 
\vspace{-2mm}
\subsection{Masked Model Design}
To properly implement masked language modeling for time-series data we use span masking, first ued in ~\cite{joshi-etal-2020-spanbert} for text and then adapted by ~\cite{10.1145/3485730.3485937} for sensor data, to prevent the trivialization of the MLM objective. One major concern is that when randomly masking input values as typically performed in NLP, the model may be able to make accurate predictions with trivial strategies such as simply taking the closest unmasked value or the average of the unmasked values on either side, inhibiting the model’s ability to learn. Span masking solves this by masking continuous spans within the input sequence. The model then needs to learn how to predict entire spans of the input data, and the above strategies are unlikely to provide accurate results.

Formally, span masking works by sampling the lengths, $l$, of each span mask from a geometric distribution which is restricted to a maximum sequence length, $l_{max}$ of 10. We skew the geometric distribution by a factor $p$ as shown in Equation 1. When the $p$ value is lowered, the sequence lengths trend towards $l_{max}$.
\begin{equation}
P(l = k) = p(1-p)^{k-1} \epsilon(1, l_{max}) \label{eq:span_masking}
\end{equation}
We select random starting points and apply the span masks of length $l$ until we have reached the desired masking ratio. We want our final model to perform well in pretraining with the masked data, but also during supervised training on the downstream tasks. To train the model to accept both masked and unmasked data we stipulate the masking is successful with a probability of $p_m$ and that the masks are replaced by random values with a probability of $p_r$. If the input is masked, all masked values are replaced by zeros. 

\subsection{Training Objective}
Given an input sample time series $s \in \mathbb{R}^{i\times j}$, where $i$ and $j$ are positive real numbers, we define the tuple $x = (A(s), A(FFT(s)))$, where $A$ represents the adaptive pooling algorithm and FFT is the normalized Fast-Fourier-Transform function. We apply span masking as described above and then the time and frequency components of $x$ are projected separately to transformer dimension $d$ using two fully-connected neural networks $L_t$ and $L_f$:
\begin{equation}
E_i = L_t(span(A(s))) + L_f(span(A(FFT(s))))
\end{equation}
We then pass our input embeddings $E_i$ to our stacked transformer encoder model $M$ to obtain output embeddings $E_o$. To reconstruct the masked portions of each input signal, $T_p$ and $F_p$, we apply two fully-connected neural networks. The reconstruction loss is then given by:
\begin{equation}
L = \frac{1}{n} \sum_{i=q_1}^{q_n} (T_{p_i} - T_{m_i})^2 + \frac{1}{n} \sum_{i=q_1}^{q_n} (F_{p_i} - F_{m_i})^2
\end{equation}
where $T_m$ and $F_m$ are the original inputs, and $q_i$ $\in$ $Q$ are the masked indices of the sequence.

\section{Experiment Setup}
\paragraph{Datasets.}
\texttt{ADAPT} is trained on 162 different datasets used for classification, their properties are summarized in Table \ref{tab:datasets_train}. Of the total 162 datasets, 158 come from the UEA and UCR time-series archive~\cite{dau2019ucr}. We also include time-series datasets SleepEEG~\cite{867928}, FD-A~\cite{lessfda}, HAR~\cite{misc_human_activity_recognition_using_smartphones_240} and ECG~\cite{Clifford2017}. The data was split by ~\cite{zhang2022selfsupervised}. All data is normalized the data at the channel level by subtracting the mean and dividing by the standard deviation. In total our training datasets consist of almost 550,000 samples.




We test \texttt{ADAPT} on several popular classification benchmarks within the time series domain. These datasets are described in Table \ref{tab:datasets_test}. Each dataset has a small amount of data for training and validation in order to challenge the limits of each self-supervised learning method. 
\paragraph{Baselines.} We compare our model with the state-of-the-art models: TS-SD~\cite{Ts-sd}, TS2vec~\cite{Yue_Wang_Duan_Yang_Huang_Tong_Xu_2022}, CLOCS~\cite{Kiyasseh2020CLOCSCL}, MIXING-UP~\cite{10.1016/j.patrec.2022.02.007}, TS-TCC~\cite{ijcai2021p324}, SimCLR~\cite{tang2021exploring} and TF-C~\cite{zhang2022selfsupervised}. Each of these methods is pre-trained on the SleepEEG dataset as it presents complicated temporal dynamics and the largest dataset for pre-training. We report baselines results from ~\cite{zhang2022selfsupervised} for comparison. Each model is fine-tuned five times with varying random seeds and we take the average across all trails. 

To disentangle the performance gains due to improvements in our pre-training algorithm via the time and frequency span masking, compared to the benefits of mixed-batch, many-to-one pre-training, we train a model called \texttt{ADAPT(EEG)}. This model is identical to ADAPT, including the use of adaptive pooling to the same input dimensions, except that it is trained on only the SleepEEG dataset (as with the state-of-the-art baselines).



\begin{table}[t]
\centering
\small
\setlength{\tabcolsep}{4pt}
\renewcommand{\arraystretch}{0.95}

\begin{minipage}[t]{0.49\columnwidth}
\centering
\caption{Testing datasets and their respective data size.}
\label{tab:datasets_test}
\begin{tabular}{@{}l r l r r@{}}
\toprule
\textbf{Dataset} & \textbf{Len.} & \textbf{Train/Val/Test} & \textbf{Ch.} & \textbf{Cls.} \\
\midrule
Epilepsy & 178    & 60 / 20 / 11\,420   & 1 & 2 \\
FD-B     & 5\,120 & 60 / 21 / 13\,559   & 1 & 3 \\
Gesture  & 315    & 320 / 120 / 120     & 3 & 8 \\
EMG      & 1\,500 & 122 / 41 / 41       & 1 & 3 \\
\bottomrule
\end{tabular}
\end{minipage}\hfill
\begin{minipage}[t]{0.49\columnwidth}
\centering
\caption{Datasets used during pre-training.}
\label{tab:datasets_train}
\begin{tabular}{@{}l r r r r@{}}
\toprule
\textbf{Dataset} & \textbf{Len.} & \textbf{Samples} & \textbf{Ch.} & \textbf{Cls.} \\
\midrule
UEA/UCR  & 8--17\,984  & 12--30\,000  & 1--1345 & 2--60 \\
SleepEEG & 200         & 371\,055     & 1       & 5 \\
FD-A     & 5\,120      & 8\,184       & 1       & 3 \\
HAR      & 128         & 10\,299      & 9       & 6 \\
ECG      & 1\,500      & 43\,673      & 1       & 4 \\
\bottomrule
\end{tabular}
\end{minipage}
\end{table}

\paragraph{Implementation Details.}
Our model is trained for 1000 epochs with a batch size of 1024 on two NVIDIA A40 GPU's. We use a base learning rate of $5e-4$ following a cosine loss schedule. Our chosen optimizer is AdamW~\cite{loshchilov2018decoupled} using $\beta_1 = 0.9, \beta_2 =0.999$ and warmup for 40 epochs. We clip the gradients at $max\char`_norm = 1$. 

Our core architecture consists of six stacked transformer encoders \cite{NIPS2017_3f5ee243} with a hidden dimensional size of 128. For the span-masking algorithm applied to the time and frequency representations we use $l_{max} = 10$, $p = 0.2$, $p_m = 0.8$ and $p_r = 0.2$.
\section{Experimental Results }

\begin{table}[t]
\caption{Overall classification accuracy of \texttt{ADAPT} compared to other state-of-the-art pre-trained models for time-series classification. Baseline implementations are from \cite{zhang2022selfsupervised} and are pretrained on the SleepEEG dataset. We pre-train two models, ADAPT(EEG) which is only pre-trained on the SleepEEG datasets, and \texttt{ADAPT} which is trained on all of the pretraining datasets in Section 5.2. The subscripts are standard deviations.}
\label{table:FullClassificationResults}
\centering
\scriptsize
\setlength{\tabcolsep}{2.5pt}
\renewcommand{\arraystretch}{0.95}

\begin{tabular}{@{}p{0.495\columnwidth}@{\hspace{0.01\columnwidth}}p{0.495\columnwidth}@{}}
\toprule
\multicolumn{1}{c}{\textbf{Dataset: Epilepsy}} & \multicolumn{1}{c}{\textbf{Dataset: FD-B}} \\
\midrule

\begin{minipage}[t]{\linewidth}\centering
\begin{tabular}{@{}lrrrr@{}}
\textbf{Model} & \textbf{Acc} & \textbf{Prec} & \textbf{Rec} & \textbf{F1} \\
\midrule
TS-SD      & 89.5$_{5.2}$ & 80.2$_{22.4}$ & 76.5$_{14.9}$ & 77.7$_{18.6}$ \\
TS2vec     & 94.0$_{0.4}$ & 90.6$_{1.2}$  & 90.4$_{1.2}$  & 90.5$_{0.7}$  \\
CLOCS      & 95.1$_{0.3}$ & 93.0$_{0.7}$  & 91.3$_{1.7}$  & 92.1$_{0.7}$  \\
Mixing-up  & 80.2$_{0.0}$ & 40.1$_{0.0}$  & 50.0$_{0.0}$  & 44.5$_{0.0}$  \\
TS-TCC     & 92.5$_{1.0}$ & 94.5$_{0.5}$  & 81.8$_{2.6}$  & 86.3$_{2.2}$  \\
SimCLR     & 90.7$_{3.4}$ & 92.2$_{1.7}$  & 78.6$_{10.7}$ & 81.8$_{10.0}$ \\
TF-C       & 94.9$_{2.5}$ & 94.6$_{1.1}$  & 89.1$_{2.2}$  & 91.5$_{5.3}$  \\
ADAPT(EEG) & 88.6$_{3.5}$ & 82.2$_{4.2}$  & 91.1$_{1.6}$  & 84.8$_{3.9}$  \\
ADAPT      & 93.6$_{2.9}$ & 90.1$_{4.9}$  & 91.7$_{0.9}$  & 88.5$_{4.6}$  \\
\end{tabular}
\end{minipage}
&
\begin{minipage}[t]{\linewidth}\centering
\begin{tabular}{@{}lrrrr@{}}
\textbf{Model} & \textbf{Acc} & \textbf{Prec} & \textbf{Rec} & \textbf{F1} \\
\midrule
TS-SD      & 55.7$_{2.1}$ & 57.1$_{5.4}$  & 60.5$_{2.7}$  & 57.0$_{3.3}$  \\
TS2vec     & 47.9$_{1.1}$ & 43.4$_{0.9}$  & 48.4$_{2.0}$  & 43.9$_{1.1}$  \\
CLOCS      & 49.3$_{3.1}$ & 48.2$_{3.2}$  & 58.7$_{3.9}$  & 47.5$_{4.9}$  \\
Mixing-up  & 67.9$_{2.5}$ & 71.5$_{3.4}$  & 76.1$_{2.0}$  & 72.7$_{2.3}$  \\
TS-TCC     & 55.0$_{2.2}$ & 52.8$_{2.9}$  & 64.0$_{1.8}$  & 54.2$_{3.4}$  \\
SimCLR     & 49.2$_{4.4}$ & 54.5$_{10.2}$ & 47.6$_{8.9}$  & 42.2$_{11.4}$ \\
TF-C       & 69.4$_{2.3}$ & 75.6$_{3.5}$  & 72.0$_{2.6}$  & 74.9$_{2.7}$  \\
ADAPT(EEG) & 97.3$_{2.8}$ & 98.2$_{1.8}$  & 98.0$_{2.0}$  & 98.0$_{2.0}$  \\
ADAPT      & 91.2$_{4.2}$ & 91.0$_{4.1}$  & 91.8$_{4.2}$  & 88.8$_{5.7}$  \\
\end{tabular}
\end{minipage}
\\

\midrule
\multicolumn{1}{c}{\textbf{Dataset: Gesture}} & \multicolumn{1}{c}{\textbf{Dataset: EMG}} \\
\midrule

\begin{minipage}[t]{\linewidth}\centering
\begin{tabular}{@{}lrrrr@{}}
\textbf{Model} & \textbf{Acc} & \textbf{Prec} & \textbf{Rec} & \textbf{F1} \\
\midrule
TS-SD      & 69.2$_{4.4}$ & 67.0$_{4.7}$  & 68.7$_{4.9}$  & 66.6$_{4.4}$  \\
TS2vec     & 69.2$_{3.3}$ & 65.5$_{3.6}$  & 68.5$_{3.5}$  & 65.7$_{3.9}$  \\
CLOCS      & 44.3$_{5.2}$ & 42.4$_{7.9}$  & 44.3$_{5.2}$  & 40.1$_{6.0}$  \\
Mixing-up  & 69.3$_{2.3}$ & 67.2$_{2.3}$  & 69.3$_{2.3}$  & 65.0$_{3.1}$  \\
TS-TCC     & 71.9$_{3.5}$ & 71.4$_{3.5}$  & 71.7$_{3.7}$  & 69.8$_{3.6}$  \\
SimCLR     & 48.0$_{5.9}$ & 59.5$_{16.2}$ & 54.1$_{19.5}$ & 49.6$_{18.7}$ \\
TF-C       & 76.4$_{2.0}$ & 77.3$_{3.6}$  & 74.3$_{2.7}$  & 75.7$_{3.1}$  \\
ADAPT(EEG) & 72.5$_{1.2}$ & 70.8$_{0.8}$  & 72.5$_{1.2}$  & 70.7$_{0.7}$  \\
ADAPT      & 77.0$_{2.5}$ & 74.9$_{3.7}$  & 77.0$_{2.5}$  & 75.1$_{2.9}$  \\
\end{tabular}
\end{minipage}
&
\begin{minipage}[t]{\linewidth}\centering
\begin{tabular}{@{}lrrrr@{}}
\textbf{Model} & \textbf{Acc} & \textbf{Prec} & \textbf{Rec} & \textbf{F1} \\
\midrule
TS-SD      & 46.1$_{0.0}$ & 15.5$_{0.0}$  & 33.3$_{0.0}$  & 21.1$_{0.0}$  \\
TS2vec     & 78.5$_{3.2}$ & 80.4$_{7.5}$  & 67.9$_{4.0}$  & 67.7$_{5.0}$  \\
CLOCS      & 69.9$_{3.2}$ & 53.1$_{7.5}$  & 53.5$_{2.9}$  & 51.4$_{4.1}$  \\
Mixing-up  & 30.2$_{5.3}$ & 11.0$_{1.3}$  & 25.8$_{4.6}$  & 15.4$_{2.0}$  \\
TS-TCC     & 78.9$_{1.9}$ & 58.5$_{9.7}$  & 63.1$_{9.9}$  & 59.0$_{9.5}$  \\
SimCLR     & 61.5$_{5.8}$ & 53.6$_{17.2}$ & 49.9$_{12.1}$ & 47.1$_{14.9}$ \\
TF-C       & 81.7$_{2.9}$ & 72.7$_{3.5}$  & 81.6$_{2.9}$  & 76.8$_{3.1}$  \\
ADAPT(EEG) & 96.6$_{4.5}$ & 94.4$_{6.1}$  & 97.3$_{3.6}$  & 95.0$_{6.2}$  \\
ADAPT      & 98.5$_{1.2}$ & 96.7$_{2.7}$  & 98.8$_{1.0}$  & 97.6$_{2.0}$  \\
\end{tabular}
\end{minipage}
\\

\bottomrule
\end{tabular}
\end{table}

\subsection{Classification Performance}
For comparison we train two versions of the \texttt{ADAPT} architecture, one that is trained on all of the datasets in Table \ref{tab:datasets_train} and one that is only trained on the SleepEEG dataset as with the other baselines. The performances of \texttt{ADAPT} and other models on the classification benchmarks are shown in Table \ref{table:FullClassificationResults}. As noted in previous work ~\cite{zhang2022selfsupervised}, pretraining time-series models on many datasets can degrade the quality of the model. We compare the performance of our model to other state-of-the-art models on in-domain pretraining on the Epilepsy dataset. 
For in-domain classification \texttt{ADAPT} is competitive with the other state of the art base-lines. Interestingly, \texttt{ADAPT} outperforms \texttt{ADAPT(EEG)} which indicates that our pretraining architecture benefits from mixed batch pretraining and that including datasets from outside the target domain can increase the performance of our model.

As previously stated, the focus within the time-series domain has been on training models which can generalize to unseen domains. This importance has been exacerbated by the limitations of pre-trained time-series models to singular datasets. \texttt{ADAPT} outperforms other state-of-the-art models across the generalization datasets on average, particularly within the FD-B and EMG datasets where both \texttt{ADAPT} models greatly improve the state-of-the-art performance. Interestingly the \texttt{ADAPT(EEG)} model outperforms \texttt{ADAPT} on the FD-B benchmark. Considering that \texttt{ADAPT(EEG)} provided worse performance than \texttt{ADAPT} on the Epilepsy (in-domain) benchmark, it is not clear why it it achieves close to perfect classification accuracy on this benchmark. We believe this may be a result of vastly increasing the diversity of the training data between the two models without proportionally increasing the amount of training data, resulting in some instabilities in model performance. \texttt{ADAPT} outperforms \texttt{ADAPT(EEG)} on all the other generalization benchmarks.

A key contribution is that contrary to what has been observed in previous literature, \texttt{ADAPT} \textbf{does not rapidly regrade performance in the many-to-one scenario}. \texttt{ADAPT} sets several new state-of-the-art performances for domain adaptation and generalization. It is also easily amenable to any model architecture. Our results justify both the effectiveness of mixed-batch pretraining in certain key downstream generalization scenarios and also the effectiveness of our time-frequency masking strategy during pretraining.

\subsection{Performance Compared to Dataset Properties}
\begin{figure}[h]
  \centering
  \includegraphics[width=0.4\columnwidth]{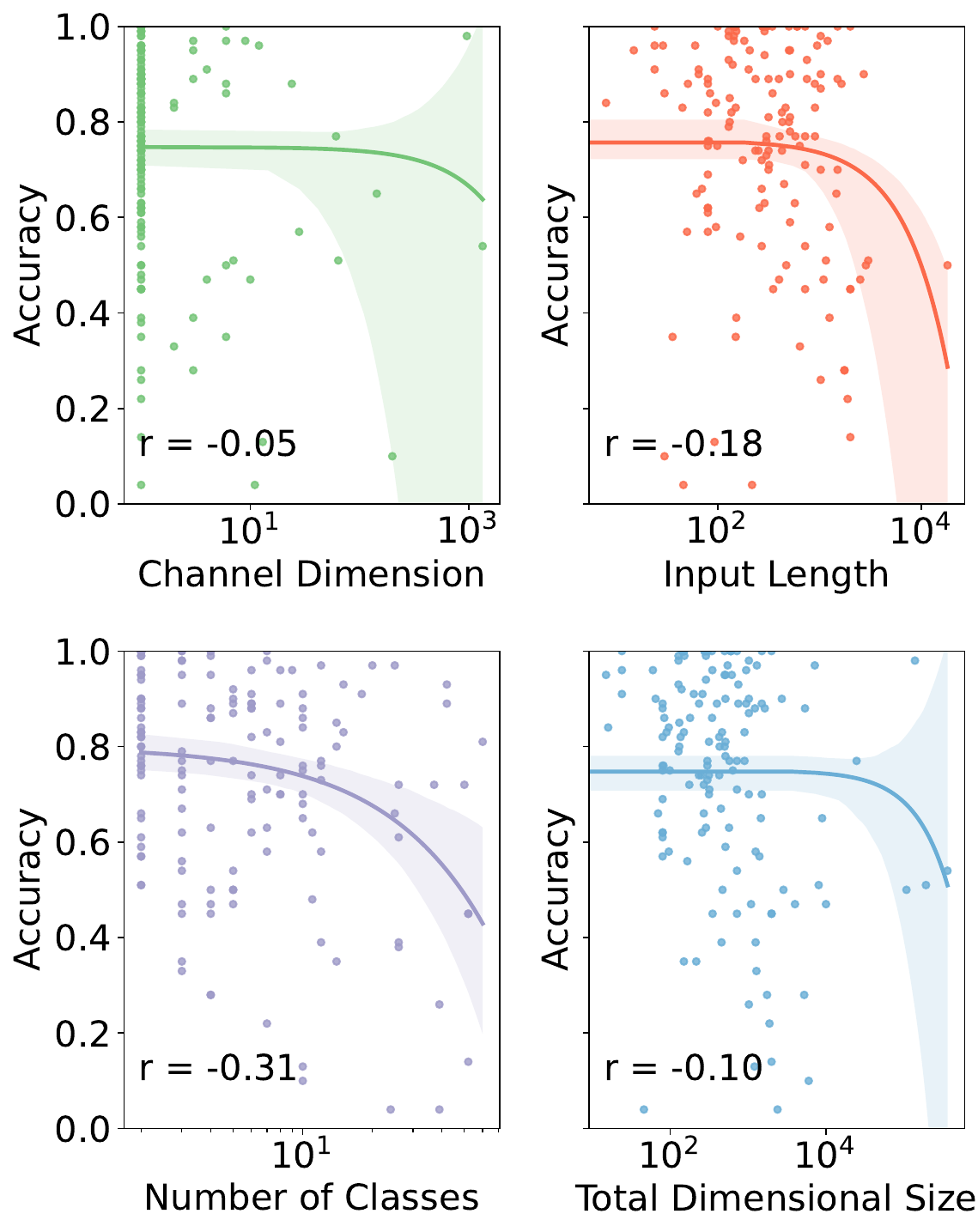}
  \caption{Model accuracy on UCR and UEA datasets compared to basic dataset properties. The ``Length'' and ``Channel Dimension'' corresponds to the dimensionality of the data or the total length of the input and number of channels at each time period. ``Number of Classes'' corresponds to the number of classes for classification in the downstream dataset, and ``Total Dimensional Size'' refers to length of the input multiplied by the number of channel dimensions. Overall, we can see that the performance is not strongly linked to the physical properties of the datasets. The number of classes of a dataset could represent an increase in dataset complexity and the model has a weak negative correlation as expected.}
  \label{fig:ACC_vs_prop}
\end{figure}

So far we have demonstrated that \texttt{ADAPT} is an effective pretraining method with allows many-to-one pretraining without degrading the accuracy of the pre-trained model as noted in previous works ~\cite{zhang2022selfsupervised}. We compare the model performance across the UCR and UEA archives with several key dataset properties in order to elucidate some insights on the effectiveness of adaptive input training. We compare the accuracy with the number of channel dimensions, input length, number of classes and the ratio of training and testing data available during finetuning in Table \ref{fig:ACC_vs_prop}. Along with each comparison we show the Pearson correlation between accuracy and the corresponding dataset property. Interestingly, the correlations between the length and channel complexity of the input and the accuracy of the model are extremely weak. We further explore the performance of \texttt{ADAPT} with respect to total dimensional size. Total dimensional size is given by dataset length multiplied by the number of channels. We chose a model embedding size of (256,32). Any dataset that is significantly below this total dimensional size will be up-sampled and those above this threshold will be down-sampled in order to meet these requirements. These results indicate that the model accuracy is not strongly correlated with the up-sampling or down-sampling of the dataset. This leads us to the conclusion that the model accuracy is more dependant on the specific task type rather than a specific dataset property. The accuracy was slightly correlated to the number of distinct classes within the dataset which futher supports this claim. These findings are highly encouraging as it shows that \texttt{ADAPT} can be a general embedding strategy for time-series data and can be used universally independent of dataset properties. For details regarding the model accuracy on each of the 158 datasets  in the UCR and UEA archives.


\subsection{Representation Visualization for Classification}
\begin{figure}[!h]
  \centering
  \includegraphics[width=0.4\columnwidth]{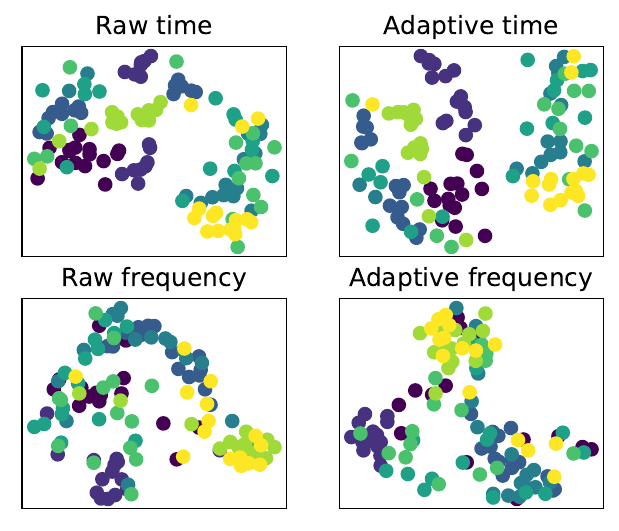}
  \caption{T-SNE visualisation of the time and frequency inputs before and after adaptive pooling on the Gesture dataset. We note that there is no discernible drop in the quality of inputs after they are transformed. The relative groupings and distributions between the 8 different classes remain largely intact.}
  \label{fig:TSNE}
\end{figure}

One potential concern is that adaptive pooling could significantly distort the input space. We use T-SNE~\cite{JMLR:v9:vandermaaten08a} to visualize the change in put in both the time and frequency domain caused by adaptive pooling. Figure \ref{fig:TSNE} shows the T-SNE analysis for the Gesture dataset, chosen for its diverse number of classes and channels. While there are small differences between the raw and adaptive inputs, the clustering of the classes and their relation to one-another remains largely intact. 

\section{Conclusion}
In this paper we introduce \texttt{ADAPT} to investigate many-to-one pre-training strategies for time-series analysis. Our adaptive pooling embedding strategy allows for training a time-series model using over 150 datasets and establishing a new state of the art on different classification benchmarks. We demonstrate that adaptive pooling can be applied as a time-series embedding strategies which can handle data over a wide range of physical properties, input lengths and modalities, as well as dataset complexity with diverse classes. \texttt{ADAPT} only changes the physical properties of the dataset and does not make any special considerations for the modality, showing that it was the diversity of the physical dataset properties (input length and channel dimension) that have prevent time-series models from learning quality representations in the many-to-one training scenario. We hope our work will help enable the creation of capable foundation models for the time-series and sensor data and bring the advancements in these areas in line with other domains.

\section*{Acknowledgements}
We would like to acknowledge the support and funding from Ingenuity Labs Research Institute at Queen's University and Ingenuity Labs Seed Funding. This research was enabled in part by support provided by Calcul Québec and the Digital Research Alliance of Canada. This Research is partially supported by NSERC Discovery Grants.

\printbibliography[heading=subbibintoc]

@inproceedings{
zhang2022selfsupervised,
title={Self-Supervised Contrastive Pre-Training For Time Series via Time-Frequency Consistency},
author={Xiang Zhang and Ziyuan Zhao and Theodoros Tsiligkaridis and Marinka Zitnik},
booktitle={Advances in Neural Information Processing Systems},
editor={Alice H. Oh and Alekh Agarwal and Danielle Belgrave and Kyunghyun Cho},
year={2022},
url={https://openreview.net/forum?id=OJ4mMfGKLN}
}

@inproceedings{
ozyurt2023contrastive,
title={Contrastive Learning for Unsupervised Domain Adaptation of Time Series},
author={Yilmazcan Ozyurt and Stefan Feuerriegel and Ce Zhang},
booktitle={The Eleventh International Conference on Learning Representations },
year={2023},
url={https://openreview.net/forum?id=xPkJYRsQGM}
}

@inproceedings{
woo2022cost,
title={Co{ST}: Contrastive Learning of Disentangled Seasonal-Trend Representations for Time Series Forecasting},
author={Gerald Woo and Chenghao Liu and Doyen Sahoo and Akshat Kumar and Steven Hoi},
booktitle={International Conference on Learning Representations},
year={2022},
url={https://openreview.net/forum?id=PilZY3omXV2}
}

@misc{Stanford_CRFM_2021,
  author       = {{Stanford CRFM}},
  title        = {Stanford Center for Research on Foundation Models (CRFM)},
  year         = {2021},
  howpublished = {Online},
  url          = {https://crfm.stanford.edu/},
  note         = {Accessed: 2024-01-30}
}

@article{amoatey2024effects,
  title={The effects of diurnal temperature range on mortality and emergency department presentations in Victoria state of Australia: A time-series analysis},
  author={Amoatey, Patrick and Osborne, Nicholas J and Darssan, Darsy and Xu, Zhiwei and Doan, Quang-Van and Phung, Dung},
  journal={Environmental Research},
  volume={240},
  pages={117397},
  year={2024},
  publisher={Elsevier}
}

@Article{app132212374,
AUTHOR = {Kashpruk, Nataliia and Piskor-Ignatowicz, Cezary and Baranowski, Jerzy},
TITLE = {Time Series Prediction in Industry 4.0: A Comprehensive Review and Prospects for Future Advancements},
JOURNAL = {Applied Sciences},
VOLUME = {13},
YEAR = {2023},
NUMBER = {22},
ARTICLE-NUMBER = {12374},
URL = {https://www.mdpi.com/2076-3417/13/22/12374},
ISSN = {2076-3417},
DOI = {10.3390/app132212374}
}

@misc{radford2021learning,
      title={Learning Transferable Visual Models From Natural Language Supervision}, 
      author={Alec Radford and Jong Wook Kim and Chris Hallacy and Aditya Ramesh and Gabriel Goh and Sandhini Agarwal and Girish Sastry and Amanda Askell and Pamela Mishkin and Jack Clark and Gretchen Krueger and Ilya Sutskever},
      year={2021},
      eprint={2103.00020},
      archivePrefix={arXiv},
      primaryClass={cs.CV}
}

@inproceedings{
li2023time,
title={Time Series as Images: Vision Transformer for Irregularly Sampled Time Series},
author={Zekun Li and Shiyang Li and Xifeng Yan},
booktitle={Thirty-seventh Conference on Neural Information Processing Systems},
year={2023},
url={https://openreview.net/forum?id=ZmeAoWQqe0}
}

@ARTICLE{867928,

  author={Kemp, B. and Zwinderman, A.H. and Tuk, B. and Kamphuisen, H.A.C. and Oberye, J.J.L.},

  journal={IEEE Transactions on Biomedical Engineering}, 

  title={Analysis of a sleep-dependent neuronal feedback loop: the slow-wave microcontinuity of the EEG}, 

  year={2000},

  volume={47},

  number={9},

  pages={1185-1194},

  doi={10.1109/10.867928}}

@misc{misc_human_activity_recognition_using_smartphones_240,
  author       = {Reyes-Ortiz,Jorge and Anguita,Davide and Ghio,Alessandro and Oneto,Luca and Parra,Xavier},
  title        = {{Human Activity Recognition Using Smartphones}},
  year         = {2012},
  howpublished = {UCI Machine Learning Repository},
  note         = {{DOI}: https://doi.org/10.24432/C54S4K}
}

@inproceedings{10.5555/3495724.3496768,
author = {Baevski, Alexei and Zhou, Henry and Mohamed, Abdelrahman and Auli, Michael},
title = {Wav2vec 2.0: A Framework for Self-Supervised Learning of Speech Representations},
year = {2020},
isbn = {9781713829546},
publisher = {Curran Associates Inc.},
address = {Red Hook, NY, USA},
booktitle = {Proceedings of the 34th International Conference on Neural Information Processing Systems},
articleno = {1044},
numpages = {12},
location = {Vancouver, BC, Canada},
series = {NIPS'20}
}

@misc{zhou2023fits,
      title={One Fits All:Power General Time Series Analysis by Pretrained LM}, 
      author={Tian Zhou and PeiSong Niu and Xue Wang and Liang Sun and Rong Jin},
      year={2023},
      eprint={2302.11939},
      archivePrefix={arXiv},
      primaryClass={cs.LG}
}

@inproceedings{
wu2023timesnet,
title={TimesNet: Temporal 2D-Variation Modeling for General Time Series Analysis},
author={Haixu Wu and Tengge Hu and Yong Liu and Hang Zhou and Jianmin Wang and Mingsheng Long},
booktitle={The Eleventh International Conference on Learning Representations },
year={2023},
url={https://openreview.net/forum?id=ju_Uqw384Oq}
}

@misc{chang2024llm4ts,
      title={LLM4TS: Aligning Pre-Trained LLMs as Data-Efficient Time-Series Forecasters}, 
      author={Ching Chang and Wei-Yao Wang and Wen-Chih Peng and Tien-Fu Chen},
      year={2024},
      eprint={2308.08469},
      archivePrefix={arXiv},
      primaryClass={cs.LG}
}

@misc{jiang2024mixtral,
      title={Mixtral of Experts}, 
      author={Albert Q. Jiang and Alexandre Sablayrolles and Antoine Roux and Arthur Mensch and Blanche Savary and Chris Bamford and Devendra Singh Chaplot and Diego de las Casas and Emma Bou Hanna and Florian Bressand and Gianna Lengyel and Guillaume Bour and Guillaume Lample and Lélio Renard Lavaud and Lucile Saulnier and Marie-Anne Lachaux and Pierre Stock and Sandeep Subramanian and Sophia Yang and Szymon Antoniak and Teven Le Scao and Théophile Gervet and Thibaut Lavril and Thomas Wang and Timothée Lacroix and William El Sayed},
      year={2024},
      eprint={2401.04088},
      archivePrefix={arXiv},
      primaryClass={cs.LG}
}

@misc{dau2019ucr,
      title={The UCR Time Series Archive}, 
      author={Hoang Anh Dau and Anthony Bagnall and Kaveh Kamgar and Chin-Chia Michael Yeh and Yan Zhu and Shaghayegh Gharghabi and Chotirat Ann Ratanamahatana and Eamonn Keogh},
      year={2019},
      eprint={1810.07758},
      archivePrefix={arXiv},
      primaryClass={cs.LG}
}

@inproceedings{ijcai2021p324,
  title     = {Time-Series Representation Learning via Temporal and Contextual Contrasting},
  author    = {Eldele, Emadeldeen and Ragab, Mohamed and Chen, Zhenghua and Wu, Min and Kwoh, Chee Keong and Li, Xiaoli and Guan, Cuntai},
  booktitle = {Proceedings of the Thirtieth International Joint Conference on
               Artificial Intelligence, {IJCAI-21}},
  publisher = {International Joint Conferences on Artificial Intelligence Organization},
  editor    = {Zhi-Hua Zhou},
  pages     = {2352--2359},
  year      = {2021},
  month     = {8},
  note      = {Main Track},
  doi       = {10.24963/ijcai.2021/324},
  url       = {https://doi.org/10.24963/ijcai.2021/324},
}

@ARTICLE{9745049,

  author={Ghazali, Mohamad Hazwan Mohd and Rahiman, Wan},

  journal={IEEE Sensors Journal}, 

  title={Vibration-Based Fault Detection in Drone Using Artificial Intelligence}, 

  year={2022},

  volume={22},

  number={9},

  pages={8439-8448},

  doi={10.1109/JSEN.2022.3163401}}

@INPROCEEDINGS{7780459,

  author={He, Kaiming and Zhang, Xiangyu and Ren, Shaoqing and Sun, Jian},

  booktitle={2016 IEEE Conference on Computer Vision and Pattern Recognition (CVPR)}, 

  title={Deep Residual Learning for Image Recognition}, 

  year={2016},

  volume={},

  number={},

  pages={770-778},

  doi={10.1109/CVPR.2016.90}}

@article{joshi-etal-2020-spanbert,
    title = "{S}pan{BERT}: Improving Pre-training by Representing and Predicting Spans",
    author = "Joshi, Mandar  and
      Chen, Danqi  and
      Liu, Yinhan  and
      Weld, Daniel S.  and
      Zettlemoyer, Luke  and
      Levy, Omer",
    journal = "Transactions of the Association for Computational Linguistics",
    volume = "8",
    year = "2020",
    address = "Cambridge, MA",
    publisher = "MIT Press",
    url = "https://aclanthology.org/2020.tacl-1.5",
    doi = "10.1162/tacl_a_00300",
    pages = "64--77",
}

@inproceedings{Kiyasseh2020CLOCSCL,
  title={CLOCS: Contrastive Learning of Cardiac Signals Across Space, Time, and Patients},
  author={Dani Kiyasseh and Tingting Zhu and David A. Clifton},
  booktitle={International Conference on Machine Learning},
  year={2020}
}

@inproceedings{lessfda,
author = {Lessmeier, Christian and Kimotho, James and Zimmer, Detmar and Sextro, Walter},
year = {2016},
month = {07},
pages = {},
title = {Condition Monitoring of Bearing Damage in Electromechanical Drive Systems by Using Motor Current Signals of Electric Motors: A Benchmark Data Set for Data-Driven Classification}
}

@article{LIU2023501,
title = {Transforming data into actionable knowledge for fault detection, diagnosis and prognosis in urban wastewater systems with AI techniques: A mini-review},
journal = {Process Safety and Environmental Protection},
volume = {172},
pages = {501-512},
year = {2023},
issn = {0957-5820},
doi = {https://doi.org/10.1016/j.psep.2023.02.043},
url = {https://www.sciencedirect.com/science/article/pii/S0957582023001428},
author = {Yiqi Liu and Pedram Ramin and Xavier Flores-Alsina and Krist V. Gernaey}
}

@inproceedings{
loshchilov2018decoupled,
title={Decoupled Weight Decay Regularization},
author={Ilya Loshchilov and Frank Hutter},
booktitle={International Conference on Learning Representations},
year={2019},
url={https://openreview.net/forum?id=Bkg6RiCqY7},
}

@misc{openai2023gpt4,
      title={GPT-4 Technical Report}, 
      author={OpenAI},
      year={2023},
      eprint={2303.08774},
      archivePrefix={arXiv},
      primaryClass={cs.CL}
}

@article{JMLR:v21:20-074,
  author  = {Colin Raffel and Noam Shazeer and Adam Roberts and Katherine Lee and Sharan Narang and Michael Matena and Yanqi Zhou and Wei Li and Peter J. Liu},
  title   = {Exploring the Limits of Transfer Learning with a Unified Text-to-Text Transformer},
  journal = {Journal of Machine Learning Research},
  year    = {2020},
  volume  = {21},
  number  = {140},
  pages   = {1--67},
  url     = {http://jmlr.org/papers/v21/20-074.html}
}

@misc{ramesh2022hierarchical,
      title={Hierarchical Text-Conditional Image Generation with CLIP Latents}, 
      author={Aditya Ramesh and Prafulla Dhariwal and Alex Nichol and Casey Chu and Mark Chen},
      year={2022},
      eprint={2204.06125},
      archivePrefix={arXiv},
      primaryClass={cs.CV}
}

@misc{reed2022generalist,
      title={A Generalist Agent}, 
      author={Scott Reed and Konrad Zolna and Emilio Parisotto and Sergio Gomez Colmenarejo and Alexander Novikov and Gabriel Barth-Maron and Mai Gimenez and Yury Sulsky and Jackie Kay and Jost Tobias Springenberg and Tom Eccles and Jake Bruce and Ali Razavi and Ashley Edwards and Nicolas Heess and Yutian Chen and Raia Hadsell and Oriol Vinyals and Mahyar Bordbar and Nando de Freitas},
      year={2022},
      eprint={2205.06175},
      archivePrefix={arXiv},
      primaryClass={cs.AI}
}

@INPROCEEDINGS{7298594,

  author={Szegedy, Christian and Wei Liu and Yangqing Jia and Sermanet, Pierre and Reed, Scott and Anguelov, Dragomir and Erhan, Dumitru and Vanhoucke, Vincent and Rabinovich, Andrew},

  booktitle={2015 IEEE Conference on Computer Vision and Pattern Recognition (CVPR)}, 

  title={Going deeper with convolutions}, 

  year={2015},

  volume={},

  number={},

  pages={1-9},

  doi={10.1109/CVPR.2015.7298594}}

@misc{tang2021exploring,
      title={Exploring Contrastive Learning in Human Activity Recognition for Healthcare}, 
      author={Chi Ian Tang and Ignacio Perez-Pozuelo and Dimitris Spathis and Cecilia Mascolo},
      year={2021},
      eprint={2011.11542},
      archivePrefix={arXiv},
      primaryClass={cs.LG}
}

@misc{thoppilan2022lamda,
      title={LaMDA: Language Models for Dialog Applications}, 
      author={Romal Thoppilan and Daniel De Freitas and Jamie Hall and Noam Shazeer and Apoorv Kulshreshtha and Heng-Tze Cheng and Alicia Jin and Taylor Bos and Leslie Baker and Yu Du and YaGuang Li and Hongrae Lee and Huaixiu Steven Zheng and Amin Ghafouri and Marcelo Menegali and Yanping Huang and Maxim Krikun and Dmitry Lepikhin and James Qin and Dehao Chen and Yuanzhong Xu and Zhifeng Chen and Adam Roberts and Maarten Bosma and Vincent Zhao and Yanqi Zhou and Chung-Ching Chang and Igor Krivokon and Will Rusch and Marc Pickett and Pranesh Srinivasan and Laichee Man and Kathleen Meier-Hellstern and Meredith Ringel Morris and Tulsee Doshi and Renelito Delos Santos and Toju Duke and Johnny Soraker and Ben Zevenbergen and Vinodkumar Prabhakaran and Mark Diaz and Ben Hutchinson and Kristen Olson and Alejandra Molina and Erin Hoffman-John and Josh Lee and Lora Aroyo and Ravi Rajakumar and Alena Butryna and Matthew Lamm and Viktoriya Kuzmina and Joe Fenton and Aaron Cohen and Rachel Bernstein and Ray Kurzweil and Blaise Aguera-Arcas and Claire Cui and Marian Croak and Ed Chi and Quoc Le},
      year={2022},
      eprint={2201.08239},
      archivePrefix={arXiv},
      primaryClass={cs.CL}
}

@inproceedings{
tonekaboni2021unsupervised,
title={Unsupervised Representation Learning for Time Series with Temporal Neighborhood Coding},
author={Sana Tonekaboni and Danny Eytan and Anna Goldenberg},
booktitle={International Conference on Learning Representations},
year={2021},
url={https://openreview.net/forum?id=8qDwejCuCN}
}

@misc{touvron2023llama,
      title={LLaMA: Open and Efficient Foundation Language Models}, 
      author={Hugo Touvron and Thibaut Lavril and Gautier Izacard and Xavier Martinet and Marie-Anne Lachaux and Timothée Lacroix and Baptiste Rozière and Naman Goyal and Eric Hambro and Faisal Azhar and Aurelien Rodriguez and Armand Joulin and Edouard Grave and Guillaume Lample},
      year={2023},
      eprint={2302.13971},
      archivePrefix={arXiv},
      primaryClass={cs.CL}
}

@article{JMLR:v9:vandermaaten08a,
  author  = {Laurens van der Maaten and Geoffrey Hinton},
  title   = {Visualizing Data using t-SNE},
  journal = {Journal of Machine Learning Research},
  year    = {2008},
  volume  = {9},
  number  = {86},
  pages   = {2579--2605},
  url     = {http://jmlr.org/papers/v9/vandermaaten08a.html}
}

@inproceedings{NIPS2017_3f5ee243,
 author = {Vaswani, Ashish and Shazeer, Noam and Parmar, Niki and Uszkoreit, Jakob and Jones, Llion and Gomez, Aidan N and Kaiser, \L ukasz and Polosukhin, Illia},
 booktitle = {Advances in Neural Information Processing Systems},
 editor = {I. Guyon and U. Von Luxburg and S. Bengio and H. Wallach and R. Fergus and S. Vishwanathan and R. Garnett},
 pages = {},
 publisher = {Curran Associates, Inc.},
 title = {Attention is All you Need},
 url = {https://proceedings.neurips.cc/paper_files/paper/2017/file/3f5ee243547dee91fbd053c1c4a845aa-Paper.pdf},
 volume = {30},
 year = {2017}
}

@inproceedings{
wagh2022evaluating,
title={Evaluating Latent Space Robustness and Uncertainty of {EEG}-{ML} Models under Realistic Distribution Shifts},
author={Neeraj Wagh and Jionghao Wei and Samarth Rawal and Brent M. Berry and Yogatheesan Varatharajah},
booktitle={Advances in Neural Information Processing Systems},
editor={Alice H. Oh and Alekh Agarwal and Danielle Belgrave and Kyunghyun Cho},
year={2022},
url={https://openreview.net/forum?id=KRk0lBRPpOC}
}

@article{10.1016/j.patrec.2022.02.007,
author = {Wickstr\o{}m, Kristoffer and Kampffmeyer, Michael and Mikalsen, Karl \O{}yvind and Jenssen, Robert},
title = {Mixing up Contrastive Learning: Self-Supervised Representation Learning for Time Series},
year = {2022},
issue_date = {Mar 2022},
publisher = {Elsevier Science Inc.},
address = {USA},
volume = {155},
number = {C},
issn = {0167-8655},
url = {https://doi.org/10.1016/j.patrec.2022.02.007},
doi = {10.1016/j.patrec.2022.02.007},
journal = {Pattern Recogn. Lett.},
month = {mar},
pages = {54–61},
numpages = {8}
}

@inproceedings{Ts-sd,
author = {Shi, Jin and Ye, Wenwen and Qin, Zheng},
year = {2021},
month = {07},
pages = {1-8},
title = {Self-Supervised Pre-training for Time Series Classification},
doi = {10.1109/IJCNN52387.2021.9533426}
}

@inproceedings{10.1145/3485730.3485937,
    author = {Xu, Huatao and Zhou, Pengfei and Tan, Rui and Li, Mo and Shen, Guobin},
    title = {LIMU-BERT: Unleashing the Potential of Unlabeled Data for IMU Sensing Applications},
    year = {2021},
    isbn = {9781450390972},
    publisher = {Association for Computing Machinery},
    address = {New York, NY, USA},
    url = {https://doi.org/10.1145/3485730.3485937},
    doi = {10.1145/3485730.3485937},
    booktitle = {Proceedings of the 19th ACM Conference on Embedded Networked Sensor Systems},
    pages = {220–233},
    numpages = {14},
    location = {Coimbra, Portugal},
    series = {SenSys '21}
    }

@article{Yue_Wang_Duan_Yang_Huang_Tong_Xu_2022, title={TS2Vec: Towards Universal Representation of Time Series}, volume={36}, url={https://ojs.aaai.org/index.php/AAAI/article/view/20881}, DOI={10.1609/aaai.v36i8.20881}, abstractNote={This paper presents TS2Vec, a universal framework for learning representations of time series in an arbitrary semantic level. Unlike existing methods, TS2Vec performs contrastive learning in a hierarchical way over augmented context views, which enables a robust contextual representation for each timestamp. Furthermore, to obtain the representation of an arbitrary sub-sequence in the time series, we can apply a simple aggregation over the representations of corresponding timestamps. We conduct extensive experiments on time series classification tasks to evaluate the quality of time series representations. As a result, TS2Vec achieves significant improvement over existing SOTAs of unsupervised time series representation on 125 UCR datasets and 29 UEA datasets. The learned timestamp-level representations also achieve superior results in time series forecasting and anomaly detection tasks. A linear regression trained on top of the learned representations outperforms previous SOTAs of time series forecasting. Furthermore, we present a simple way to apply the learned representations for unsupervised anomaly detection, which establishes SOTA results in the literature. The source code is publicly available at https://github.com/yuezhihan/ts2vec.}, number={8}, journal={Proceedings of the AAAI Conference on Artificial Intelligence}, author={Yue, Zhihan and Wang, Yujing and Duan, Juanyong and Yang, Tianmeng and Huang, Congrui and Tong, Yunhai and Xu, Bixiong}, year={2022}, month={Jun.}, pages={8980-8987} }

@misc{zhou2023comprehensive,
      title={A Comprehensive Survey on Pretrained Foundation Models: A History from BERT to ChatGPT}, 
      author={Ce Zhou and Qian Li and Chen Li and Jun Yu and Yixin Liu and Guangjing Wang and Kai Zhang and Cheng Ji and Qiben Yan and Lifang He and Hao Peng and Jianxin Li and Jia Wu and Ziwei Liu and Pengtao Xie and Caiming Xiong and Jian Pei and Philip S. Yu and Lichao Sun},
      year={2023},
      eprint={2302.09419},
      archivePrefix={arXiv},
      primaryClass={cs.AI}
}

@inproceedings{Clifford2017,
   author = {Gari Clifford and Chengyu Liu and Benjamin Moody and Li-wei Lehman and Ikaro Silva and Qiao Li and Alistair Johnson and Roger Mark},
   doi = {10.22489/CinC.2017.065-469},
   month = {9},
   title = {AF Classification from a Short Single Lead ECG Recording: the Physionet Computing in Cardiology Challenge 2017},
   year = {2017},
}

@article{Amankulova2023,
   author = {Khilola Amankulova and Nizom Farmonov and László Mucsi},
   doi = {10.1016/j.atech.2022.100098},
   issn = {27723755},
   journal = {Smart Agricultural Technology},
   month = {2},
   pages = {100098},
   title = {Time-series analysis of Sentinel-2 satellite images for sunflower yield estimation},
   volume = {3},
   year = {2023},
}

@article{Acosta2022,
   author = {Julián N. Acosta and Guido J. Falcone and Pranav Rajpurkar and Eric J. Topol},
   doi = {10.1038/s41591-022-01981-2},
   issn = {1078-8956},
   issue = {9},
   journal = {Nature Medicine},
   month = {9},
   pages = {1773-1784},
   title = {Multimodal biomedical AI},
   volume = {28},
   year = {2022},
}

@article{10.1145/3502289,
author = {Cao, Longbing},
title = {AI in Finance: Challenges, Techniques, and Opportunities},
year = {2022},
issue_date = {March 2023},
publisher = {Association for Computing Machinery},
address = {New York, NY, USA},
volume = {55},
number = {3},
issn = {0360-0300},
url = {https://doi.org/10.1145/3502289},
doi = {10.1145/3502289},
journal = {ACM Comput. Surv.},
month = {feb},
articleno = {64},
numpages = {38}
}

\end{document}


\onecolumn

\appendices

\section{Detailed Dataset Description}
Below we summarize each dataset used during both pretraining and finetuning/testing phases of our experiments:
\begin{itemize}
\item \textbf{UCR/UEA} is an archive of 158 time-series classification datasets. Of these, 128 are univariate in that it has one channel dimension and 30 are multivariate with a range of channel dimensions, the rough dimensions are summarized in Table 1. It consists of a wide variety of data which are roughly classified into: Audio, Motion, Sensor, HAR, Device, SPECTRO, ECG, Hemodynamics, Simulated, Image, Sound, Meg and Misc. Each dataset is publicly available in its pre-processed form at ~\cite{dau2019ucr}. Refer to \textbf{APPENDIX C} for full details of all 158 datasets.

For the following datasets we use the same setup as \cite{zhang2022selfsupervised} and are appreciative for their hard work in providing the preprocessed data used within this work. This enables easy comparison of our methods to those within their work. For more information, refer to their paper and open source code.

\item\textbf{SleepEEG} contains 197 whole-night PolySomnoGraphic sleep recordings, containing sleep data collected from 82 healthy subjects. The data is univariate, recorded from a 1-lead EEG recorded at 100Hz. The data was segmented into lengths of size 200 with no overlap. There are five categories assocaited with each sample: wake(W), non-rapid eye movement (N1,N2,N3) and rapid-eye-movement (REM).

\item\textbf{Epilepsy} is a dataset consisting of single-channel EEG data from 500 subjects. The total length of the recordings for each subject was 23.6 seconds and the data was split into 11,500 samples, recorded at 178 Hz. The dataset was pre-processed to have two different classes: eyes open, eyes closed, EEG measured in healthy brain region, EEG where a tumor is located and EEG during a seizure episode. The dataset is binary classification problem wherein the first 4 classes were merged into one (no seizure).

\item\textbf{FDA/FDB} is data gathered from synchronously measured motor currents and vibrations collected from bearings under different rotation speed, load torque and radial force. There are three total classes which include outer damage, inner damage and undamaged. FD-A and FD-B denote data collected under differing operating conditions. The data was processed into lengths of 5120 with equal ratio's between classes. This data was collected at 64k Hz for 4 seconds.

\item\textbf{HAR} consists of human activity data collected from 30 healthy volunteers performing six activities. These include walking, walking upstairs, walking downstairs, sitting, standing and laying. Although data was collected from both triaxial accelerometers and gyroscopes, only the three channels of accelerometer data were used to maintain accurate comparisons to other literature. While our model can take model of varying dimensions, the other comparable models required that the channel dimensions between the Gesture and HAR datasets match for in-domain testing. There were 10,299 samples used and each was recorded at 50Hz.

\item\textbf{Gesture} Gesture contains accelerometer measurements of eight gestures that include, hand swiping left, right, up and down, hand waving counterclockwise and clockwise and hand waving in a square and right arrow pattern. The dataset consists of 440 samples and 55 samples per class. The assumed sampling frequency is 100 Hz.

\item\textbf{ECG} is taken from the 2017 PhysioNet challenge and uses a single-lead ECG for arrhythmia classification. The different types of cardiac rhythm included are: normal sinus, atrial fibrillation, alternative, or others (too much noise for classification). The data was collected at 300Hz and the atrial fibrillation class was underrepresented within the dataset. Each sample is of length 1500 which corresponds to 5 seconds of data.

\item\textbf{EMG} measures muscle responses as electrical activity and are commonly used to diagnose muscular distrophies and neuropathies. This dataset consists of a single channel EMG from the tibialis anterior muscle of three subjects, one of which is healthy, one with neuropathy and one with myopathy. Each recording is sampled at 4k Hz and the total input size is 1500.
   
\end{itemize}
\section{Hyperparameters for Classification}
We tune the hyperparameters of \texttt{ADAPT} for each downstream classification dataset. For easy reproduction of our results we summarize the hyperparameters in Table \ref{table:hyperparams}. The LC denotes that we froze all of the model paramters except for the final linear layer during finetuning. Each experiment is repeated 5 times with a different random seed. Our transformer based model has a hidden dimension of 128 and consists of 6 stacked transformer encoders.
\begin{table}[!h]
\caption{Hyperparamter summary for each task. Note for the Epilepsy dataset, we found that due to the small amount of data, better results were obtained when freezing all of the parameters except the final linear layer.}\label{table:hyperparams}
\centering
\begin{tabular}{@{}llll@{}}
\toprule
\textbf{Task} & \textbf{Batch size} & \textbf{Learning rate} & \textbf{Epochs} \\ 
\midrule
EMG & 32 & 4$\times10^{-4}$ & 5 \\ 
FD-B & 32 & 4$\times10^{-4}$ & 50 \\ 
Gesture & 32 & 1$\times10^{-4}$ & 50 \\ 
Epilepsy [LC] & 32 & 1$\times10^{-3}$ & 15 \\ 
\bottomrule
\end{tabular}
\end{table}
\pagebreak
\section{Evaluation of performance on UCR and UEA Datasets}
We finetuned our model on each of the UCR and UEA datasets to evaluate its performance across datasets of differing physical properties. As \texttt{ADAPT} only changes the physical structure of the data (its input length and dimension), we want to observe any clear biases. The performance for each dataset is listed below. We kept all hyperparameters the same and finetuned with a learning rate of 1e-4 for 200 epochs and a batch size of 32.
\setlength{\tabcolsep}{3pt}
\begin{longtable}{lrrrrrr}\toprule
                       Dataset &  Train size &  Test Size &  Dim &  Length &  Num Classes &  Accuracy \\
\midrule
\endfirsthead

\toprule
                       Dataset &  Train size &  Test Size &  Dim &  Length &  Num Classes &  Accuracy \\
\midrule
\endhead
\midrule
\multicolumn{7}{r}{{Continued on next page}} \\
\midrule
\endfoot

\bottomrule
\endlastfoot
     ArticularyWordRecognition &         275 &        300 &    9 &     144 &           25 &      0.97 \\
            AtrialFibrillation &          15 &         15 &    2 &     640 &            3 &      0.33 \\
                  BasicMotions &          40 &         40 &    6 &     100 &            4 &      1.00 \\
         CharacterTrajectories &        1422 &       1436 &    3 &     182 &           20 &      0.97 \\
                       Cricket &         108 &         72 &    6 &    1197 &           12 &      0.97 \\
                 DuckDuckGeese &          50 &         50 & 1345 &     270 &            5 &      0.54 \\
                    EigenWorms &         128 &        131 &    6 &   17984 &            5 &      0.50 \\
                      Epilepsy &         137 &        138 &    3 &     206 &            4 &      0.95 \\
          EthanolConcentration &         261 &        263 &    3 &    1751 &            4 &      0.28 \\
                         ERing &          30 &        270 &    4 &      65 &            6 &      0.91 \\
                 FaceDetection &        5890 &       3524 &  144 &      62 &            2 &      0.65 \\
               FingerMovements &         316 &        100 &   28 &      50 &            2 &      0.57 \\
         HandMovementDirection &         160 &         74 &   10 &     400 &            4 &      0.47 \\
                   Handwriting &         150 &        850 &    3 &     152 &           26 &      0.39 \\
                     Heartbeat &         204 &        205 &   61 &     405 &            2 &      0.77 \\
                InsectWingbeat &       30000 &      20000 &  200 &      30 &           10 &      0.10 \\
                JapaneseVowels &         270 &        370 &   12 &      29 &            9 &      0.96 \\
                        Libras &         180 &        180 &    2 &      45 &           15 &      0.83 \\
                          LSST &        2459 &       2466 &    6 &      36 &           14 &      0.35 \\
                  MotorImagery &         278 &        100 &   64 &    3000 &            2 &      0.51 \\
                        NATOPS &         180 &        180 &   24 &      51 &            6 &      0.88 \\
                     PenDigits &        7494 &       3498 &    2 &       8 &           10 &      0.84 \\
                       PEMS-SF &         267 &        173 &  963 &     144 &            7 &      0.98 \\
                       Phoneme &        3315 &       3353 &   11 &     217 &           39 &      0.04 \\
                  RacketSports &         151 &        152 &    6 &      30 &            4 &      0.86 \\
            SelfRegulationSCP1 &         268 &        293 &    6 &     896 &            2 &      0.88 \\
            SelfRegulationSCP2 &         200 &        180 &    7 &    1152 &            2 &      0.51 \\
            SpokenArabicDigits &        6599 &       2199 &   13 &      93 &           10 &      0.13 \\
                 StandWalkJump &          12 &         15 &    4 &    2500 &            3 &      0.47 \\
           UWaveGestureLibrary &         120 &        320 &    3 &     315 &            8 &      0.89 \\
                         Adiac &         390 &        391 &    1 &     176 &           37 &      0.72 \\
                     ArrowHead &          36 &        175 &    1 &     251 &            3 &      0.74 \\
                          Beef &          30 &         30 &    1 &     470 &            5 &      0.50 \\
                     BeetleFly &          20 &         20 &    1 &     512 &            2 &      0.95 \\
                   BirdChicken &          20 &         20 &    1 &     512 &            2 &      0.90 \\
                           Car &          60 &         60 &    1 &     577 &            4 &      0.77 \\
                           CBF &          30 &        900 &    1 &     128 &            3 &      0.98 \\
         ChlorineConcentration &         467 &       3840 &    1 &     166 &            3 &      0.56 \\
                  CinCECGTorso &          40 &       1380 &    1 &    1639 &            4 &      0.88 \\
                        Coffee &          28 &         28 &    1 &     286 &            2 &      1.00 \\
                     Computers &         250 &        250 &    1 &     720 &            2 &      1.00 \\
                      CricketX &         390 &        390 &    1 &     300 &           12 &      0.73 \\
                      CricketY &         390 &        390 &    1 &     300 &           12 &      0.77 \\
                      CricketZ &         390 &        390 &    1 &     300 &           12 &      0.76 \\
           DiatomSizeReduction &          16 &        306 &    1 &     345 &            4 &      0.86 \\
  DistalPhalanxOutlineAgeGroup &         400 &        139 &    1 &      80 &            3 &      0.75 \\
   DistalPhalanxOutlineCorrect &         600 &        276 &    1 &      80 &            2 &      0.76 \\
               DistalPhalanxTW &         400 &        139 &    1 &      80 &            6 &      0.69 \\
                   Earthquakes &         322 &        139 &    1 &     512 &            2 &      0.78 \\
                        ECG200 &         100 &        100 &    1 &      96 &            2 &      0.84 \\
                       ECG5000 &         500 &       4500 &    1 &     140 &            5 &      0.92 \\
                   ECGFiveDays &          23 &        861 &    1 &     136 &            2 &      0.82 \\
               ElectricDevices &        8926 &       7711 &    1 &      96 &            7 &      0.58 \\
                       FaceAll &         560 &       1690 &    1 &     131 &           14 &      0.80 \\
                      FaceFour &          24 &         88 &    1 &     350 &            4 &      0.45 \\
                      FacesUCR &         200 &       2050 &    1 &     131 &           14 &      0.85 \\
                    FiftyWords &         450 &        455 &    1 &     270 &           50 &      0.72 \\
                          Fish &         175 &        175 &    1 &     463 &            7 &      0.83 \\
                         FordA &        3601 &       1320 &    1 &     500 &            2 &      0.95 \\
                         FordB &        3636 &        810 &    1 &     500 &            2 &      0.80 \\
                      GunPoint &          50 &        150 &    1 &     150 &            2 &      0.83 \\
                           Ham &         109 &        105 &    1 &     431 &            2 &      0.80 \\
                  HandOutlines &        1000 &        370 &    1 &    2709 &            2 &      0.90 \\
                       Haptics &         155 &        308 &    1 &    1092 &            5 &      0.47 \\
                       Herring &          64 &         64 &    1 &     512 &            2 &      0.59 \\
                   InlineSkate &         100 &        550 &    1 &    1882 &            7 &      0.22 \\
           InsectWingbeatSound &         220 &       1980 &    1 &     256 &           11 &      0.62 \\
              ItalyPowerDemand &          67 &       1029 &    1 &      24 &            2 &      0.96 \\
        LargeKitchenAppliances &         375 &        375 &    1 &     720 &            3 &      0.77 \\
                    Lightning2 &          60 &         61 &    1 &     637 &            2 &      0.75 \\
                    Lightning7 &          70 &         73 &    1 &     319 &            7 &      0.71 \\
                        Mallat &          55 &       2345 &    1 &    1024 &            8 &      0.70 \\
                          Meat &          60 &         60 &    1 &     448 &            3 &      0.67 \\
                 MedicalImages &         381 &        760 &    1 &      99 &           10 &      0.75 \\
  MiddlePhalanxOutlineAgeGroup &         400 &        154 &    1 &      80 &            3 &      0.62 \\
   MiddlePhalanxOutlineCorrect &         600 &        291 &    1 &      80 &            2 &      0.57 \\
               MiddlePhalanxTW &         399 &        154 &    1 &      80 &            6 &      0.62 \\
                    MoteStrain &          20 &       1252 &    1 &      84 &            2 &      0.86 \\
    NonInvasiveFetalECGThorax1 &        1800 &       1965 &    1 &     750 &           42 &      0.89 \\
    NonInvasiveFetalECGThorax2 &        1800 &       1965 &    1 &     750 &           42 &      0.93 \\
                      OliveOil &          30 &         30 &    1 &     570 &            4 &      0.63 \\
                       OSULeaf &         200 &        242 &    1 &     427 &            6 &      0.74 \\
      PhalangesOutlinesCorrect &        1800 &        858 &    1 &      80 &            2 &      0.61 \\
                       Phoneme &         214 &       1896 &    1 &    1024 &           39 &      0.26 \\
                         Plane &         105 &        105 &    1 &     144 &            7 &      1.00 \\
ProximalPhalanxOutlineAgeGroup &         400 &        205 &    1 &      80 &            3 &      0.89 \\
 ProximalPhalanxOutlineCorrect &         600 &        291 &    1 &      80 &            2 &      0.88 \\
             ProximalPhalanxTW &         400 &        205 &    1 &      80 &            6 &      0.82 \\
          RefrigerationDevices &         375 &        375 &    1 &     720 &            3 &      0.54 \\
                    ScreenType &         375 &        375 &    1 &     720 &            3 &      0.45 \\
                   ShapeletSim &          20 &        180 &    1 &     500 &            2 &      1.00 \\
                     ShapesAll &         600 &        600 &    1 &     512 &           60 &      0.81 \\
        SmallKitchenAppliances &         375 &        375 &    1 &     720 &            3 &      0.71 \\
         SonyAIBORobotSurface1 &          20 &        601 &    1 &      70 &            2 &      0.66 \\
         SonyAIBORobotSurface2 &          27 &        953 &    1 &      65 &            2 &      0.90 \\
               StarLightCurves &        1000 &       8236 &    1 &    1024 &            3 &      0.98 \\
                    Strawberry &         613 &        370 &    1 &     235 &            2 &      0.89 \\
                   SwedishLeaf &         500 &        625 &    1 &     128 &           15 &      0.93 \\
                       Symbols &          25 &        995 &    1 &     398 &            6 &      0.89 \\
              SyntheticControl &         300 &        300 &    1 &      60 &            6 &      0.96 \\
              ToeSegmentation1 &          40 &        228 &    1 &     277 &            2 &      1.00 \\
              ToeSegmentation2 &          36 &        130 &    1 &     343 &            2 &      1.00 \\
                         Trace &         100 &        100 &    1 &     275 &            4 &      1.00 \\
                    TwoLeadECG &          23 &       1139 &    1 &      82 &            2 &      0.76 \\
                   TwoPatterns &        1000 &       4000 &    1 &     128 &            4 &      0.99 \\
        UWaveGestureLibraryAll &         896 &       3582 &    1 &     945 &            8 &      0.96 \\
          UWaveGestureLibraryX &         896 &       3582 &    1 &     315 &            8 &      0.81 \\
          UWaveGestureLibraryY &         896 &       3582 &    1 &     315 &            8 &      0.70 \\
          UWaveGestureLibraryZ &         896 &       3582 &    1 &     315 &            8 &      0.74 \\
                         Wafer &        1000 &       6164 &    1 &     152 &            2 &      0.99 \\
                          Wine &          57 &         54 &    1 &     234 &            2 &      0.74 \\
                  WordSynonyms &         267 &        638 &    1 &     270 &           25 &      0.66 \\
                         Worms &         181 &         77 &    1 &     900 &            5 &      0.77 \\
                 WormsTwoClass &         181 &         77 &    1 &     900 &            2 &      0.83 \\
                          Yoga &         300 &       3000 &    1 &     426 &            2 &      0.82 \\
                         ACSF1 &         100 &        100 &    1 &    1460 &           10 &      0.65 \\
            AllGestureWiimoteX &         300 &        700 &    1 &       0 &           10 &      0.68 \\
            AllGestureWiimoteY &         300 &        700 &    1 &       0 &           10 &      0.76 \\
            AllGestureWiimoteZ &         300 &        700 &    1 &       0 &           10 &      0.70 \\
                           BME &          30 &        150 &    1 &     128 &            3 &      0.79 \\
                     Chinatown &          20 &        343 &    1 &      24 &            2 &      1.00 \\
                          Crop &        7200 &      16800 &    1 &      46 &           24 &      0.04 \\
                 DodgerLoopDay &          78 &         80 &    1 &     288 &            7 &      0.63 \\
                DodgerLoopGame &          20 &        138 &    1 &     288 &            2 &      0.94 \\
             DodgerLoopWeekend &          20 &        138 &    1 &     288 &            2 &      0.99 \\
           EOGHorizontalSignal &         362 &        362 &    1 &    1250 &           12 &      0.58 \\
             EOGVerticalSignal &         362 &        362 &    1 &    1250 &           12 &      0.39 \\
                  EthanolLevel &         504 &        500 &    1 &    1751 &            4 &      0.28 \\
           FreezerRegularTrain &         150 &       2850 &    1 &     301 &            2 &      1.00 \\
             FreezerSmallTrain &          28 &       2850 &    1 &     301 &            2 &      1.00 \\
                         Fungi &          18 &        186 &    1 &     201 &           18 &      0.91 \\
               GestureMidAirD1 &         208 &        130 &    1 &       0 &           26 &      0.72 \\
               GestureMidAirD2 &         208 &        130 &    1 &       0 &           26 &      0.61 \\
               GestureMidAirD3 &         208 &        130 &    1 &       0 &           26 &      0.38 \\
               GesturePebbleZ1 &         132 &        172 &    1 &       0 &            6 &      0.88 \\
               GesturePebbleZ2 &         146 &        158 &    1 &       0 &            6 &      0.89 \\
               GunPointAgeSpan &         135 &        316 &    1 &     150 &            2 &      1.00 \\
      GunPointMaleVersusFemale &         135 &        316 &    1 &     150 &            2 &      1.00 \\
        GunPointOldVersusYoung &         136 &        315 &    1 &     150 &            2 &      1.00 \\
                   HouseTwenty &          40 &        119 &    1 &    2000 &            2 &      1.00 \\
         InsectEPGRegularTrain &          62 &        249 &    1 &     601 &            3 &      1.00 \\
           InsectEPGSmallTrain &          17 &        249 &    1 &     601 &            3 &      1.00 \\
           MelbournePedestrian &        1194 &       2439 &    1 &      24 &           10 &      0.91 \\
       MixedShapesRegularTrain &         500 &       2425 &    1 &    1024 &            5 &      0.90 \\
         MixedShapesSmallTrain &         100 &       2425 &    1 &    1024 &            5 &      0.87 \\
         PickupGestureWiimoteZ &          50 &         50 &    1 &       0 &           10 &      0.86 \\
             PigAirwayPressure &         104 &        208 &    1 &    2000 &           52 &      0.14 \\
                PigArtPressure &         104 &        208 &    1 &    2000 &           52 &      0.45 \\
                        PigCVP &         104 &        208 &    1 &    2000 &           52 &      0.45 \\
                         PLAID &         537 &        537 &    1 &       0 &           11 &      0.48 \\
                     PowerCons &         180 &        180 &    1 &     144 &            2 &      1.00 \\
                          Rock &          20 &         50 &    1 &    2844 &            4 &      0.50 \\
             SemgHandGenderCh2 &         300 &        600 &    1 &    1500 &            2 &      1.00 \\
           SemgHandMovementCh2 &         450 &        450 &    1 &    1500 &            6 &      0.70 \\
            SemgHandSubjectCh2 &         450 &        450 &    1 &    1500 &            5 &      0.89 \\
          ShakeGestureWiimoteZ &          50 &         50 &    1 &       0 &           10 &      0.88 \\
                SmoothSubspace &         150 &        150 &    1 &      15 &            3 &      0.95 \\
                           UMD &          36 &        144 &    1 &     150 &            3 &      0.35 \\
\end{longtable}

A summary of the datasets and their properties is shown below including the range of results. We did not perform any dataset specific hyperparameter tuning, the overall accuracy could be improved if we tuned for each dataset although with significant computation costs. Note that a length of 0 denotes that the input length of that dataset varies.
\begin{longtable}{lrrrrrr}
\toprule
  Score &  Train\_size &  Test\_Size &  Dim &  Length &  Num\_Classes &  Accuracy \\
\midrule
\endfirsthead

\toprule
  Score &  Train\_size &  Test\_Size &  Dim &  Length &  Num\_Classes &  Accuracy \\
\midrule
\endhead
\midrule
\multicolumn{7}{r}{{Continued on next page}} \\
\midrule
\endfoot

\bottomrule
\endlastfoot
average &         769 &       1086 &   20 &     601 &            9 &      0.75 \\
    min &          12 &         15 &    1 &       0 &            2 &      0.04 \\
    max &       30000 &      20000 & 1345 &   17984 &           60 &      1.00 \\
\end{longtable}

\section{T-SNE Settings}
We conduct T-SNE analysis on the the raw model embeddings before and after adaptive pooling, as well as the model embeddings after pretraining. Our dimensional space for analysis is two dimensional and we use a default perplexity of 30 and the T-SNE algorithm is run for 100 iterations. 
\section{Fundamental Challenges}

\subsection{Domain Shift}
Domain shift is a common challenge in time series modeling and can be described as a drop in performance caused by differences in distribution of the training dataset and downstream task. Consider a time series model $M$ trained on a training dataset $D_{train} = {X_1, X_2, ..., X_n}$, where each $X_i$ represents a time series sample. We assume that the training dataset $D_{train}$ is sampled from a distribution $p(X)$.
When we apply the trained model $M$ to a downstream dataset $D_{test} = {X_1', X_2', ..., X_m'}$, where each $X_i'$ represents a time series sample, we assume that the test dataset $D_{test}$ is sampled from a distribution $p(X')$ which may differ from the distribution $p(X)$ of the training dataset. This difference in underlying distributions can lead to a phenomenon called domain shift. We can express the change in model performance as a function of loss $L$:
\begin{equation}
\text{Domain Shift} = L(M(D_{test}), Y_{test}) -L(M(D_{train}), Y_{train})
\end{equation}
Where $M(D_{test})$ is the model's predictions on the test dataset, $Y_{test}$ is the corresponding ground truth labels, $M(D_{train})$ is the model's predictions on the training dataset, and $Y_{train}$ is the corresponding ground truth labels.The domain shift phenomenon can occur due to various reasons such as differences in sampling methods, data acquisition,small differences in recording equipment, data pre-processing, and data labeling between the training and test datasets. In recent works, self-supervised learning methods have shown promising results in alleviating the effects of domain shift in time series models. It is important to note that domain shift occurs within time-series of the same modality, for example, between two different datasets containing EEG readings~\cite{wagh2022evaluating}.

\subsection{Domain Adaptation}
The major focus of recent works has been to improve the domain adaptation qualities of pre-trained models within the time series domain. Let $X$ be the input data, $Y$ be the output label, and $D$ be the data distribution. We define a training dataset $D_{train}$ to consist of $n$ independent samples from $D$, denoted as $D_{train} = {(x_i, y_i)}_{i=1}^{n}$, where $x_i \sim D$ and $y_i = f(x_i)$.
Domain generalization can be defined as the ability of a model $M$ to learn from one domain $D_1$ during self-supervised pre-training and adapt to another unseen domain $D_2$ during fine-tuning. The drop-off in accuracy during domain adaptation can be measured as the difference in loss between the test set of $D_2$ and the training set of $D_1$, denoted as $L(M(D_2), Y_{test}) - L(M(D_1), Y_{train})$, where $Y_{train}$ is the output labels corresponding to the training data in $D_1$. In general we can refer to domain adaptation is applying models to datasets which are of differing modality to those observed during pre-training. 

In order to improve a model's ability to generalize across domains, recent self-supervised learning methods have focused on improving a model's inductive biases to facilitate robustness to domain adaptation. One trivial solution to domain generalization is to train the model on more datasets, limiting the chances that the model will be applied in a domain generalization situation. However, increasing the number of pre-training datasets can actually inhibit model performance, as found in recent work by ~\cite{zhang2022selfsupervised} and until now no viable solution has been proposed enable large-scale, mixed-batch training on multiple datasets for time-series data.

\subsection{Diverse Modalities}
In the field of machine learning, there has been significant success in training large models on diverse datasets in other domains such as image (ResNet~\cite{7780459}, Inception~\cite{7298594}), text (GPT4~\cite{openai2023gpt4}, LLAMA~\cite{touvron2023llama}, LaMDA ~\cite{thoppilan2022lamda}), and controls ~\cite{reed2022generalist}. However, time-series based data presents unique challenges that have yet to be fully addressed. Specifically, there are no foundation models for time-series data as no one has successfully pre-trained a single model on a wide range of datasets and time-series modalities.
One major challenge is that time-series data has a wide range of input lengths and channel diversities, making traditional methods like padding and truncation impractical for widespread application. Padding shorter sequences leads to extreme data redundancy and model sizes, while truncating longer sequences results in significant information loss. Another solution is to use separate embedding models for each modality type, but this requires training an excessive number of embedding models (over 150+ for our model).
To address these challenges, a strategy for embedding data that is input size agnostic is needed for practical and efficient computation. In the next section, we introduce a training strategy called ``Adaptive Input Training,'' which enables mixed batch training for time-series data.
Mathematically, we can represent the challenges of time-series data as the need to handle input sequences of varying lengths and diverse modalities. Specifically, let $S$ denote the set of all time-series data and let s $\in$ $S$ be a particular time-series. Then, the input sequence for $s$ can be represented as $x_1$, $x_2$, ..., $x_T$, where $T$ is the length of the sequence. Additionally, each time-series may have multiple modalities, such as sensor data and audio data, which further complicates the input representation.

To train a single model on a wide range of time-series data, we need to develop a strategy for embedding the input sequence $s$ that is independent of its length T and modality type. The goal is to transform the input sequence into a fixed-length embedding vector that can be fed into a downstream model for further processing. The challenge is to design an embedding function that can handle sequences of varying length and diverse modalities in an efficient and effective manner. To perform powerful mixed batch training, this embedding needs to be applied during either data-loading or pre-processing which means that there can be no trainable parameters.

\section{Formal Definitions}
We have created a more detailed definition on the implementation of Adaptive Pooling, noise addition and the span masking algorithm.

Given an input sample time series \(s \in \mathbb{R}^{i \times j}\), where \(i\) and \(j\) are the dimensions of the series, we first apply the Fast Fourier Transform (FFT) to obtain the frequency components, denoted as \(\text{FFT}(s)\). Next, we apply an adaptive pooling algorithm \(A\) separately to both \(s\) and \(\text{FFT}(s)\), resulting in \(A(s)\) and \(A(\text{FFT}(s))\), respectively.

After adaptive pooling, we add Gaussian noise \(\mathcal{N}_{i' \times j'}(\mu, \sigma^2)\) to both \(A(s)\) and \(A(\text{FFT}(s))\), yielding noisy versions of the pooled time and frequency components, denoted as \(s_{\text{n}}\) and \(\text{FFT}(s)_{\text{n}}\), respectively:
\[s_{\text{n}} = A(s) + \mathcal{N}_{i' \times j'}(\mu, \sigma^2)\]
\[\text{FFT}(s)_{\text{n}} = A(\text{FFT}(s)) + \mathcal{N}_{i' \times j'}(\mu, \sigma^2)\]

Following the addition of noise, we apply span masking to both \(s_{\text{n}}\) and \(\text{FFT}(s)_{\text{n}}\). The time and frequency components of the masked and $n$ signals are then separately projected to the transformer dimension \(d\) using two fully-connected neural networks, \(L_t\) and \(L_f\):
\[E_i = L_t(\text{span}(s_{\text{n}})) + L_f(\text{span}(\text{FFT}(s)_{\text{n}}))\]

\subsubsection*{Adaptive Pooling}

Finally we wanted to shed more light on how adaptive pooling is implemented under the hood, particularly with the dynamic computation of kernels to enable us to pool to any output dimension, regardless of the the shape of the input tensor.

Given a 1-dimensional input tensor $T$ of length $N$, and the desired output dimension $M$, the result of adaptive pooling can be computed as follows:
    Let $K_i$ represent the kernel for the $i^{th}$ output element, where $i = 1, 2, \ldots, M$. Each $K_i$ is a subset of indexes into $T$ determined by the adaptive pooling algorithm:
    \begin{equation}
        K_i = \{ start(i), \ldots, end(i) \}
    \end{equation}
    where $start(i)$ and $end(i)$ are calculated using the formulas:
    \begin{equation}
        start(i) = \left\lfloor \frac{(i - 1) \times N}{M} \right\rfloor
    \end{equation}
    \begin{equation}
        end(i) = \left\lceil \frac{i \times N}{M} \right\rceil
    \end{equation}
    The output of the adaptive pooling operation, $O$, is a new tensor of length $M$, where each element $O_i$ is computed as the average of the elements of $T$ indexed by $K_i$:
    \begin{equation}
        O_i = \frac{1}{|K_i|} \sum_{k \in K_i} T_k
    \end{equation}
    This process adapts the kernel sizes and strides dynamically based on the input and output dimensions, allowing for variable kernel sizes that are not necessarily integer multiples of each other.

\bibliographystyle{unsrt}  
\bibliography{ref}